\documentclass[twocolumn,letterpaper]{IEEEAerospaceCLS}
\usepackage{graphicx}
\usepackage{algorithm}
\usepackage{algorithmic}
\usepackage{amsmath}
\usepackage{amssymb}
\usepackage{makecell}
\usepackage{upgreek}
\usepackage{bm}
\usepackage{fancyhdr}
\usepackage{subcaption}
\usepackage[font=small,labelfont=bf]{caption}
\usepackage{etoolbox}

\AtBeginEnvironment{tabular}{\small}


\fancypagestyle{firstpage}{%
  \fancyhf{} 
  \renewcommand{\headrulewidth}{0pt} 
  \fancyfoot[C]{© 2025 California Institute of Technology. Government sponsorship acknowledged.} 
}

\title{A General Purpose Method for Robotic Interception of Non-Cooperative Dynamic Targets}

\author{
Tanmay P. Patel, Erica L. Tevere, Erik H. Kramer, Rudranarayan M. Mukherjee \\
Jet Propulsion Laboratory, California Institute of Technology \\
4800 Oak Grove Dr.\\
Pasadena, CA 91109 \\
erica.l.tevere@jpl.nasa.gov
\thanks{\footnotesize \copyright \quad 2025 California Institute of Technology. Government sponsorship acknowledged.}
}
\date{03 October 2025}

\begin{document}

\thispagestyle{plain}
\pagestyle{plain}

\maketitle

\thispagestyle{plain}
\pagestyle{plain}

\begin{abstract}
This paper presents a general purpose framework for autonomous, vision-based interception of dynamic, non-cooperative targets, validated across three distinct mobility platforms: an unmanned aerial vehicle (UAV), a four-wheeled ground rover, and an air-thruster spacecraft testbed. The approach relies solely on a monocular camera with fiducials for target tracking and operates entirely in the local observer frame without the need for global information. The core contribution of this work is a streamlined and general approach to autonomous interception that can be adapted across robots with varying dynamics, as well as our comprehensive study of the robot interception problem across heterogenous mobility systems under limited observability and no global localization. Our method integrates (1) an Extended Kalman Filter for relative pose estimation amid intermittent measurements, (2) a history-conditioned motion predictor for dynamic target trajectory propagation, and (3) a receding-horizon planner solving a constrained convex program in real time to ensure time-efficient and kinematically feasible interception paths. Our operating regime assumes that observability is restricted by partial fields of view, sensor dropouts, and target occlusions. Experiments are performed in these conditions and include autonomous UAV landing on dynamic targets, rover rendezvous and leader-follower tasks, and spacecraft proximity operations. Results from simulated and physical experiments demonstrate robust performance with low interception errors (both during station-keeping and upon scenario completion), high success rates under deterministic and stochastic target motion profiles, and real-time execution on embedded processors such as the Jetson Orin, VOXL2, and Raspberry Pi 5. These results highlight the framework's generalizability, robustness, and computational efficiency.
\end{abstract}

\setcounter{tocdepth}{1}
\tableofcontents

\begin{figure}[t]
    \centering
    \begin{subfigure}{0.55\linewidth}
        \centering
        \includegraphics[width=\linewidth]{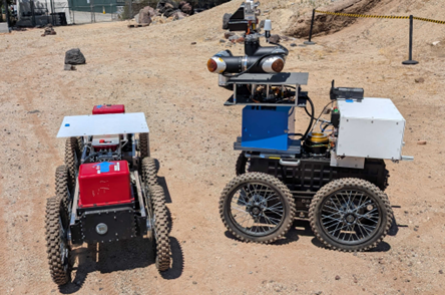}
        \caption{}
        \label{fig:rover-real}
    \end{subfigure}
    \hfill
    \begin{subfigure}{0.40\linewidth}
        \centering
        \includegraphics[width=\linewidth, trim=0 3 0 3, clip]{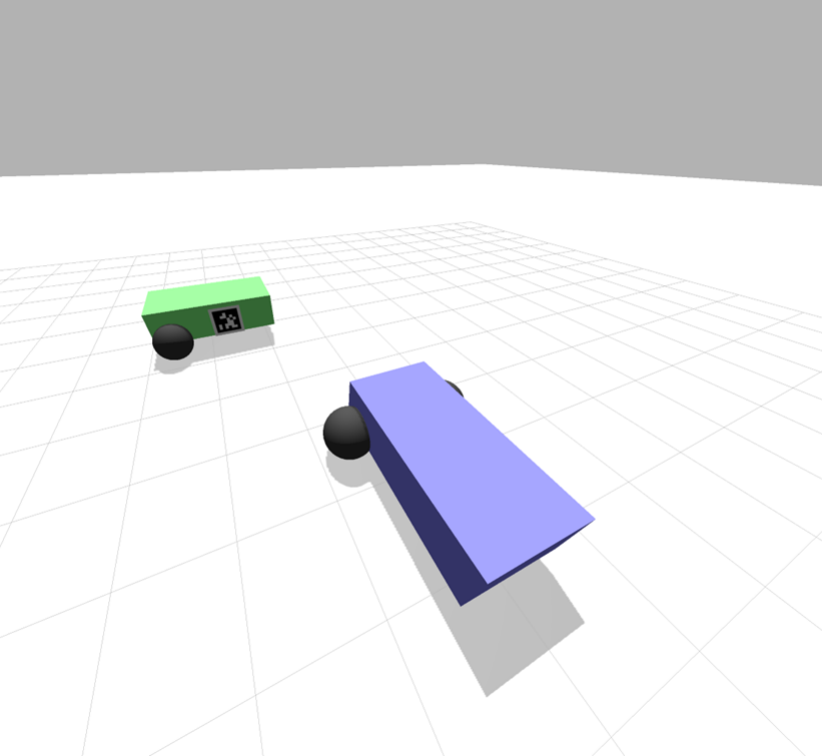}
        \caption{}
        \label{fig:rover-sim}
    \end{subfigure}

    \begin{subfigure}{0.55\linewidth}
        \centering
        \includegraphics[width=\linewidth]{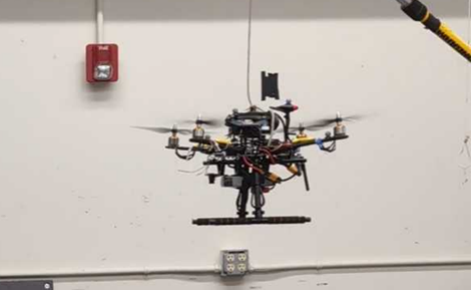}
        \caption{}
        \label{fig:uav-real}
    \end{subfigure}
    \hfill
    \begin{subfigure}{0.40\linewidth}
        \centering
        \includegraphics[width=\linewidth, trim=0 19 0 19, clip]{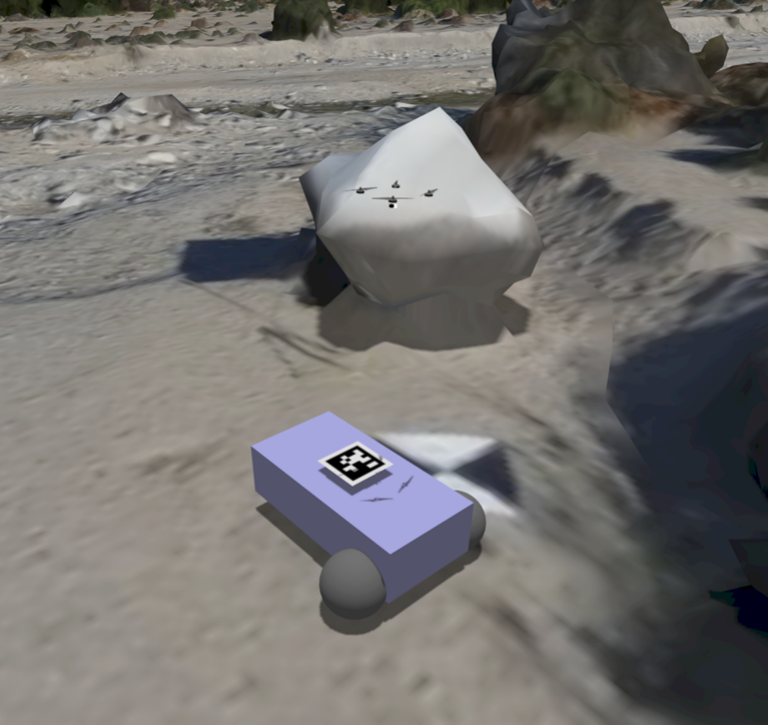}
        \caption{}
        \label{fig:uav-sim}
    \end{subfigure}

    \begin{subfigure}{0.55\linewidth}
        \centering
        \includegraphics[width=\linewidth]{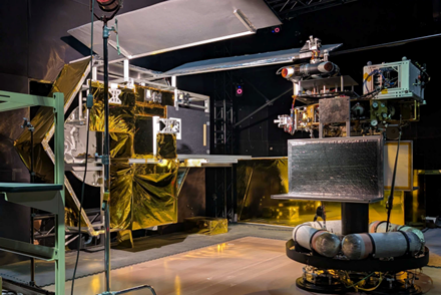}
        \caption{}
        \label{fig:gsat-real}
    \end{subfigure}
    \hfill
    \begin{subfigure}{0.40\linewidth}
        \centering
        \includegraphics[width=\linewidth, trim=91 0 91 0, clip]{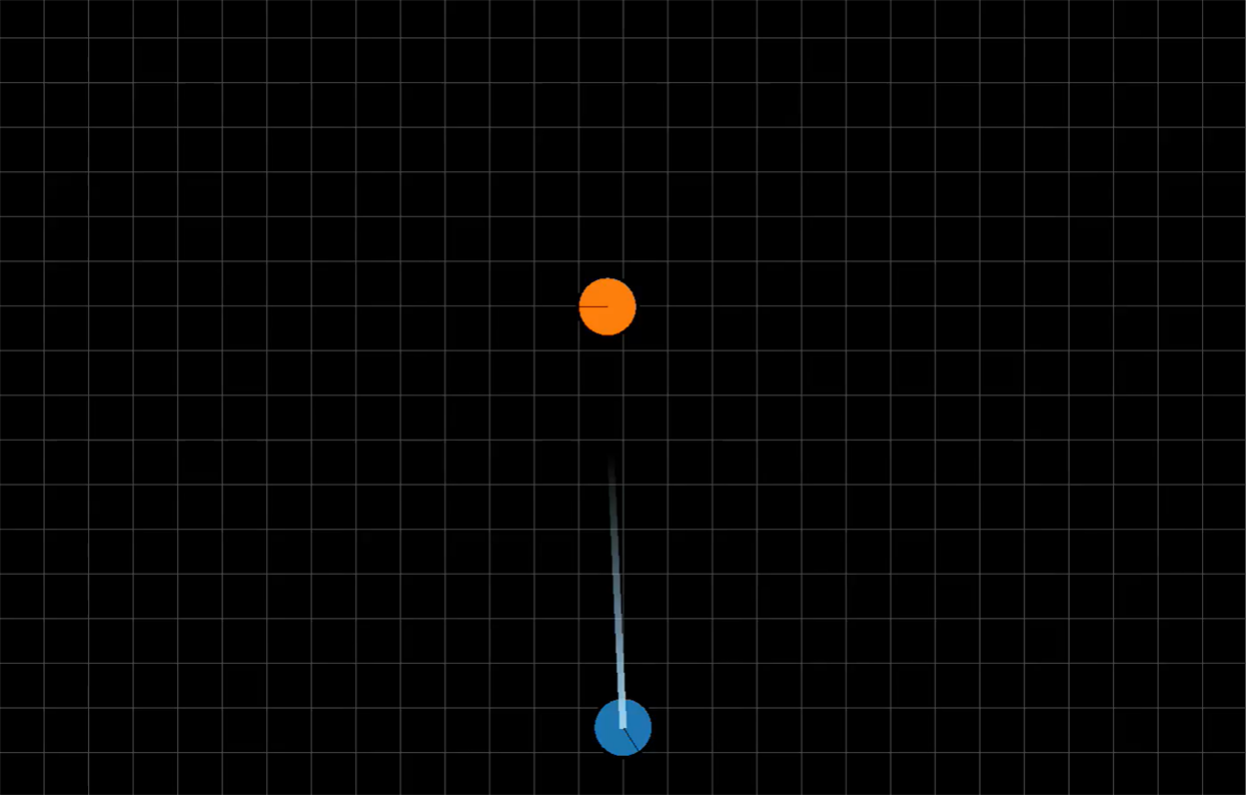}
        \caption{}
        \label{fig:gsat-sim}
    \end{subfigure}

    \caption{Examples of real-world (left) and simulated (right) scenarios that our framework can handle on three disparate robotic platforms: a rover ((a) and (b)), a UAV ((c) and (d)), and a spacecraft testbed ((e) and (f)).}
    \label{fig:real-vs-sim}
\end{figure}

\section{Introduction}

Robust autonomous interception of non-cooperative, dynamic targets is a critical capability for a wide range of space robotic systems, including rovers, multirotor UAVs, and orbital spacecraft performing proximity operations. In many planetary scenarios, global localization is unavailable or challenging \cite{ebadi2022}, sensing is limited, and observability is constrained by partial fields of view, sensor dropouts, or occlusions \cite{tunstel2002}. Under these dynamic and unknown space environments, many trajectory planning and tracking methods fail to provide sufficient adaptability and robustness. This limitation arises because they typically rely on restrictive assumptions about target dynamics, depend on specialized sensing, or incorporate platform-specific complexity that hinders their applicability across the diverse scenarios encountered in space \cite{bassil2012}.


This work introduces a general-purpose autonomy framework for real-time relative state estimation, target motion prediction, and trajectory planning to intercept non-cooperative targets under highly limited sensing conditions and partial observability. Our framework is designed to be straightforward, integrating just three modular components: an Extended Kalman Filter (EKF) for relative target pose estimation with intermittent measurements, a motion predictor that propagates a plausible target trajectory based on its observed history, and a receding-horizon trajectory optimizer that computes kinematically feasible and time-efficient interception paths while respecting platform-specific dynamic constraints. The system achieves generalizability across platforms with highly disparate dynamics and sensing configurations by operating entirely in the observer’s body-frame, requiring only a monocular camera stream, and making minimal environmental assumptions.

We carefully study the performance of our framework across multiple robotic platforms: ground rovers performing rendezvous or dynamic station-keeping, a multirotor UAV landing on a dynamic target vehicle, and thruster-based spacecraft simulators executing proximity maneuvers. In each case, the target is assumed to exhibit dynamic, stochastic, and non-cooperative---but not necessarily adversarial---motion. Since we formally address both holonomic and non-holonomic robots, our method can extend to a range of other platforms too, including satellites, autonomous cars, and home robots.

Unlike prior work, our framework is not tailored to a specific robotic platform. It makes minimal assumptions about the target, operates without any global information, and relies on a single sensing modality. Platform-dependent details such as degrees of freedom and vehicle dynamics can be incorporated through modular interfaces. This minimalist and general framework, together with a comprehensive cross-platform study of its efficacy, is the core contribution of our work, providing researchers with a robust foundation for robotic interception tasks that can be easily extended to new platforms and applications.

The remainder of this paper is structured as follows. Section \ref{sec:rel-works} reviews relevant literature. Section \ref{sec:methodology} presents our general-purpose method and discusses platform-dependent variation systematically to enable streamlined integration with varying platforms types. Section \ref{sec:results} presents results from both simulation and physical experiments. Finally, Section \ref{sec:conc} summarizes the work and outlines directions for future study.


\section{Related Works}
\pagestyle{fancy} 
\fancyhf{} 
\fancyfoot[C]{\thepage} 
\renewcommand{\headrulewidth}{0pt} 

\label{sec:rel-works}

The problem of robotic interception is well-studied and we can draw significant insights from prior work.

\subsection{Target Behaviour Prediction}

An important component of robotic interception is predicting the behaviour of the target. Qu et al. \cite{qu2022} formulate this as a multi-dimensional polynomial fit, which is simple, fast, and effective. This has the advantage of not requiring a target model, but it can be susceptible to outlier target pose estimates. Other works rely solely on an Extended Kalman Filter to propagate the target's dynamics, like \cite{falanga2017} and \cite{das2018}, but they assume a robust and accurate model of the target. In the case of \cite{falanga2017}, a constant velocity assumption for the target is used, while \cite{das2018} uses well-known projectile kinematics. We have no \emph{a priori} knowledge of the target's motion model and cannot rely on such assumptions.


Recent work has increasingly shifted toward learning-based modeling of the target, leveraging methods such as diffusion policies and foundation models. Notably, many advances in generalist navigation policies can be adapted to the interception problem. For example, \cite{zhang2024_navid} uses a single on-the-fly camera stream to navigate toward a specified goal, demonstrating an architecture capable of reasoning about target behavior and handling dynamic targets. The seminal work by Sridhar et al. \cite{sridhar2023} performs goal-conditioned navigation using diffusion policies, which can be directly applied to dynamic interception. More recently, vision-language-action (VLA) models have also gained traction. Wang et al. \cite{wang2025_trackvla} introduce a combined VLA and diffusion architecture to jointly perform target recognition and motion planning, achieving strong results with highly dynamic human targets, even under partial occlusions. 

However, these methods require large amounts of observational data and are computationally intensive, making them infeasible for applications with minimal sensing and computational power. Our primary challenge lies in achieving rapid, online performance in highly dynamic environments, a need that current learning-based methods do not specifically address.


\subsection{Trajectory Planning and Control}

The literature on trajectory planning for interception is rich. We restrict ourselves to approaches that do not consider obstacles, leaving obstacle-avoidance to future work. This obviates search-based or sampling-based methods, like A* and BIT* \cite{gammell2015},  which are not strictly necessary in the absence of obstacles.

For Ackermann robots, like our rover, the Dubins model \cite{lavalle06} and the Reeds-Shepp model \cite{lavalle06} are often used to perform interception \cite{manyam2019}. However, this method does not generalize to multiple robot classes or target profiles; for instance, \cite{manyam2019} assumes a mobile robot intercepting a target bounded to move along a circle. Bezier curves have also been used for interception trajectory planning \cite{zhang2016} but can be brittle in practice since they fail to account for the system's partial observability, providing no formal safeguard against maneuvers that cause the target to leave the observer's field of view for prolonged intervals.

Receding-horizon optimization is another widely-used strategy, in part due to its generality and flexibility \cite{paden2016}. Specifically, for robots with decoupled position and orientation degrees-of-freedom, \cite{mellinger2011_minsnap} provides a useful minimum snap and minimum acceleration formulation. This work is extended by \cite{falanga2017} to roll-out trajectories to multiple possible points of interception, after which the trajectory with minimum control effort is selected. For robots with non-holonomic dynamics, like Ackermann steering, receding-horizon optimization is also common in the form of model predictive control (MPC) \cite{babu2018}.

\section{Methodology}
\label{sec:methodology}

Our method consists of target pose estimation, target motion and intercept prediction, and observer trajectory planning. We first outline our perception setup before detailing these core components.

\subsection{Perception}

Perception is performed using OpenCV's AprilTag detection pipeline. We use this library to generate pose estimates of the target at up to 30 Hz, but as low as 5 Hz on certain hardware.

\subsection{Target Pose Estimation}

We filter the pose estimates from OpenCV using an Extended Kalman Filter (EKF) formulated entirely in the body-frame of the observer. The need for this filter stems from frequent sensor dropout and target occlusion amid off-nominal lighting conditions and motion blur.

For both the observer and target, no global information is assumed; all quantities are relative to the observer's body‑frame. This reflects the constraints of some of our platforms and the inherent difficulty of achieving real-time global state estimation in planetary robotics applications \cite{ebadi2022}.

The EKF tracks the relative state $\mathbf{x}_R = \begin{bmatrix} x_r & y_r & \theta_r \end{bmatrix}^\top$ where $(x_r, y_r)$ is the target's Cartesian position and $\theta_r$ is its heading, all expressed in the observer's body-frame.

\subsubsection{Prediction Update}
For the purposes of the EKF, the target is modeled with constant linear and angular velocity. Because the EKF state is expressed in the observer’s body-frame, the observer’s own translation and rotation also influence the state. Incorporating both effects yields a relative kinematic model suitable for a standard EKF state and covariance prediction, which is performed as per \cite{kalman1961}.

\subsubsection{Measurement Update}
Whenever a new pose estimate is available from OpenCV, we execute an EKF measurement update. Since the measurements directly observe $\mathbf{x}_R$, the update model is identity and the correction is straightforward. If no measurement is available (e.g., due to dropout), only the prediction step is performed.

\subsubsection{Target Velocity Estimation}
To perform the prediction update outlined earlier, we must estimate the target’s linear and angular velocities. Since the observer only has access to \emph{relative} pose histories, this requires \emph{cancelling} the effect of the observer’s motion. We do this by computing first differences of the relative EKF states over a sliding window of $W$ steps. This gives average relative velocities. Adding back the observer's control inputs provides estimates of the target's linear and angular velocities, which can be used to complete the prediction model.

Note that if available, we can use odometry from the observer instead of its control history to reduce drift. In either case, this step of canceling the observer's motion is critical for maintaining consistent target motion predictions in the absence of global information.

\subsection{Target Motion Prediction}

\begin{figure}[h]
    \centering
    \includegraphics[width=1\linewidth]{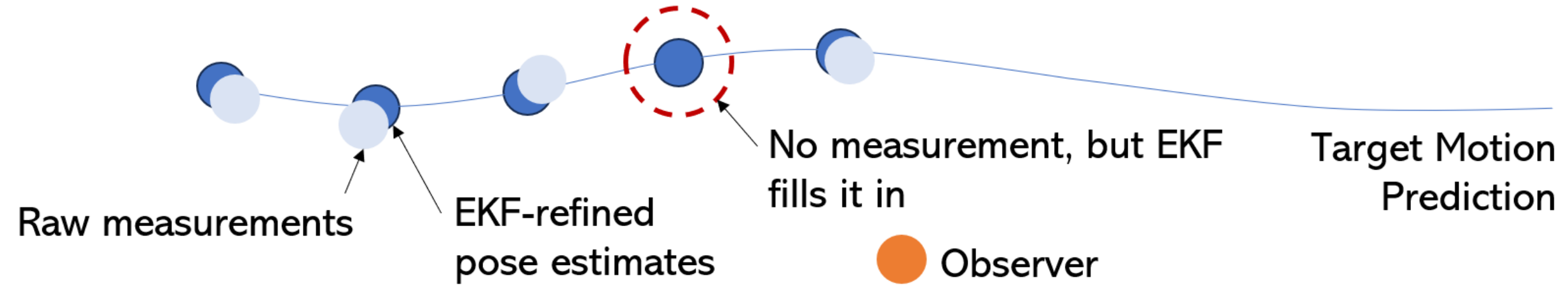}
    \caption{The predicted target trajectory is an element-wise cubic polynomial fit to the most recent EKF pose estimates. We use the EKF estimates as opposed to raw measurements to reduce sensitivity to sensor dropout and noise.}
    \label{fig:prediction}
\end{figure}

To predict the target's motion for future time steps, we draw inspiration from \cite{qu2022}. We predict future poses using polynomial regression over a sliding history of state estimates, as illustrated in Figure \ref{fig:prediction}.

Given a double-ended queue of recent target pose estimates from the EKF, $\{ \mathbf{x}_{R,-L+1}, \dots, \mathbf{x}_{R,0} \}$, with constant time spacing $\Delta t$, we fit independent $3^{\text{rd}}$-order polynomials to $x$, $y$, and $\theta$ as functions of time. We let $t_i = i \Delta t$, where $0 \le i < L$, denote the relative time indices. We construct the Vandermonde matrix as


\begin{equation}
\mathbf{T} \in \mathbb{R}^{L \times 4}, \quad T_{ij} = t_i^{3-j}, \quad 0 \le j \le 3,
\label{eq:vandermonde}
\end{equation}

and estimate the coefficient vectors as

\begin{align}
\bm{\eta}_x &= (\mathbf{T}^\top \mathbf{T})^{-1} \mathbf{T}^\top \mathbf{d}_x,
\end{align}

where $\mathbf{d}_x = [x_{r,-L+1}, \dots, x_{r,0}]^\top$. A similar computation is reproduced for $y$ and $\theta$. These vectors define time-parametric predictions for target motion, which can be expressed as

\begin{align}
\mathbf{x}(t) &\triangleq \begin{bmatrix} x(t) \\ y(t) \\ \theta(t) \end{bmatrix} = \sum_{i=0}^{n} \begin{bmatrix} \eta_{x,i} \\ \eta_{y,i} \\ \eta_{\theta,i} \end{bmatrix} \, t^{n-i},
\end{align}


where $x(t)$, $y(t)$, and $\theta(t)$ are the respective polynomials for each degree of freedom. The generality of this strategy enables its seamless application across all tested robotic platforms, requiring no customization or tuning.

\subsection{Target Intercept Prediction}

\label{sec:intercept-prediction}

\begin{figure}[h]
    \centering
    \includegraphics[width=0.85\linewidth]{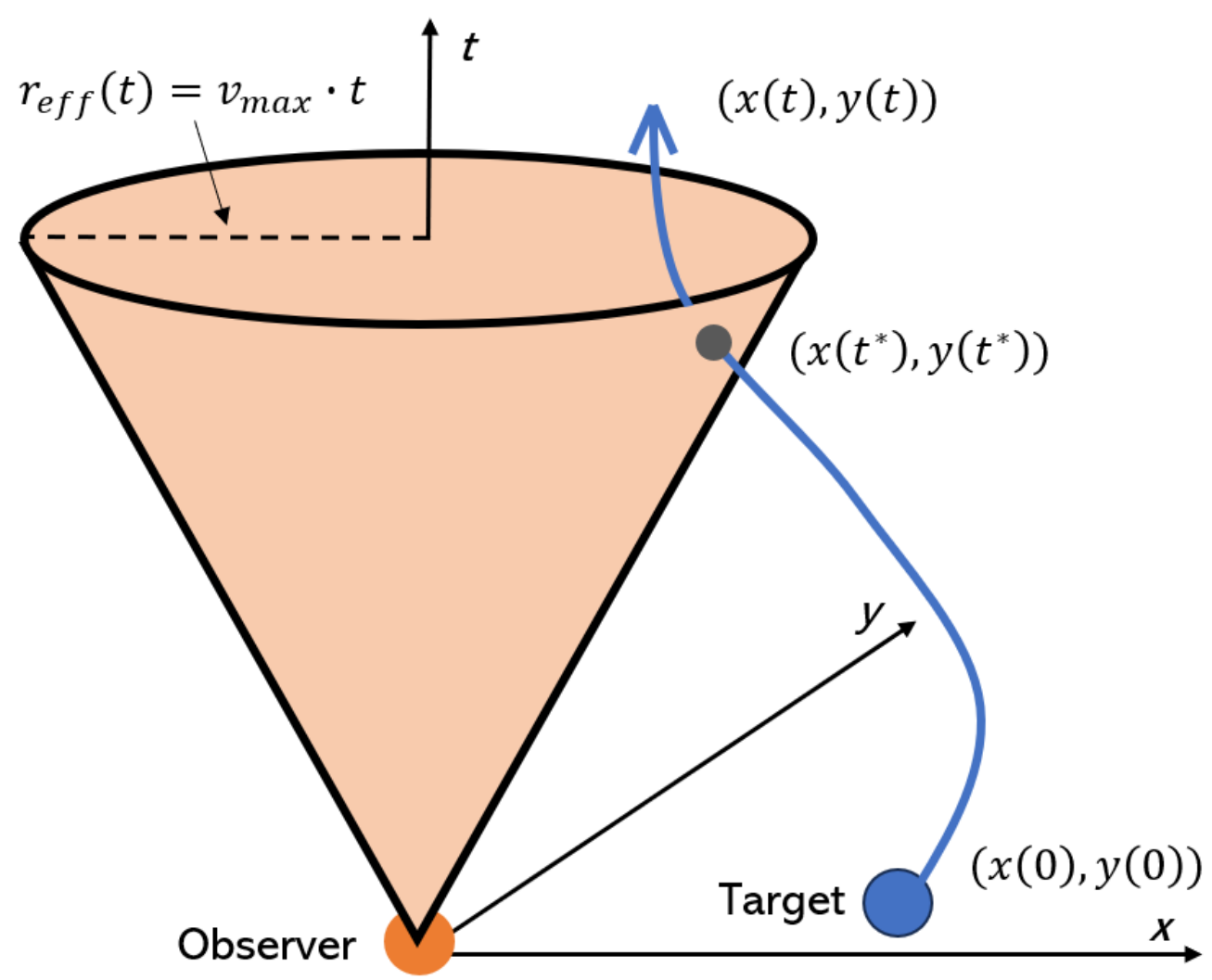}
    \caption{The orange cone denotes the total reachable area for the observer in a given amount of time, while the blue curve denotes the target's predicted motion. The earliest point at which the blue curve pierces the orange cone is the earliest point at which we can intercept and becomes the goal of our planner. Note that the cross-section of the cone depends on the platform, and in the case of the UAV, is actually a 3D cross-section owing to the additional translational degree of freedom.}
    \label{fig:reachability}
\end{figure}

We estimate the time, $t^*$, and position, $(x^*, y^*)$, at which the controlled agent (observer) can intercept the target, assuming the observer is constrained by its dynamics model and the target moves according to the profile predicted in the previous module. Platform-dependent idiosyncracies begin to manifest here. For the ground-based rover, we use Dubins reachability to predict interception since it accounts for the coupling between the observer's position and orientation. For other observers---those with uncoupled degrees of freedom---we use a simpler constant velocity reachability study. A depiction of the constant velocity case is provided in Figure \ref{fig:reachability}.

\subsubsection{Coupled Position and Orientation}

\begin{figure}[h]
    \centering
    \includegraphics[width=0.95\linewidth]{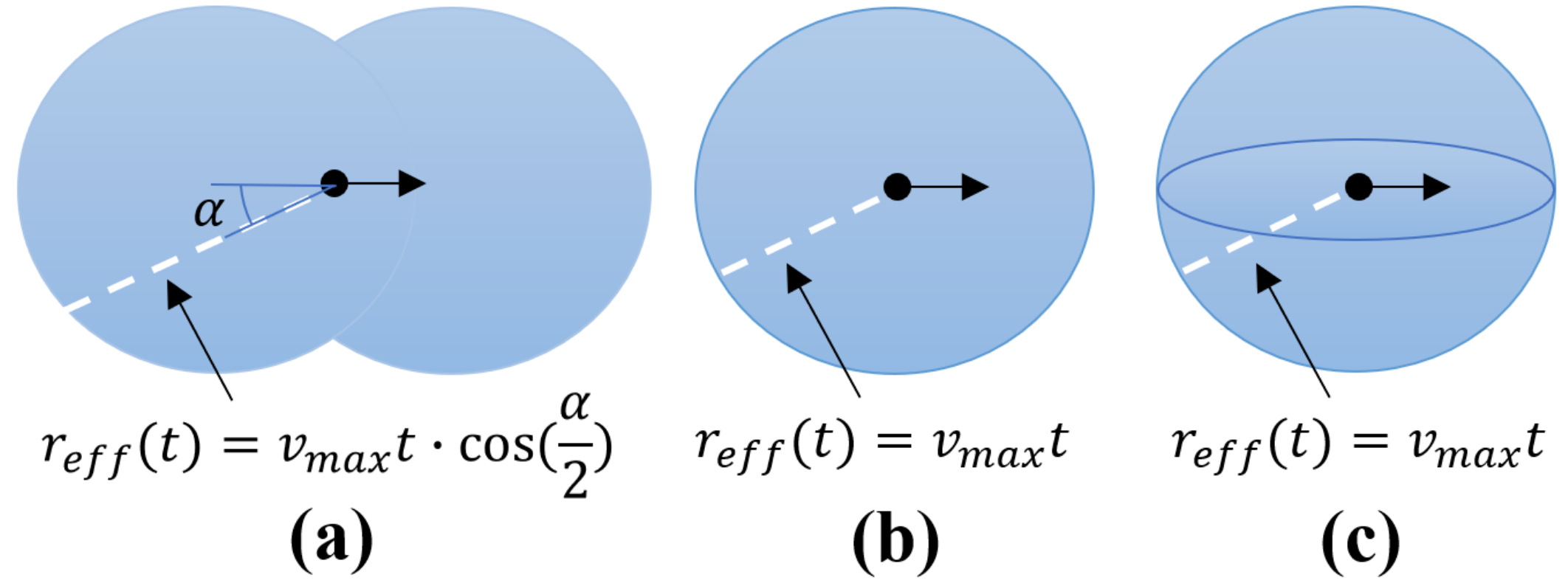}
    \caption{(a) For the rover, the reachable cross-section is peanut-shaped, with the narrow regions corresponding to states that require substantial steering to reach. (b) For the spacecraft testbed, the cross-section is circular due to its holonomic motion; this case is studied in Figure \ref{fig:reachability}. (c) For the UAV, the cross-section is spherical, reflecting the additional vertical degree of freedom. For all three cases, the black arrow indicates the observer's current heading.}
    \label{fig:cross-sections}
\end{figure}

We model the rover, whose positional dynamics depend on heading, as a Dubins car constrained by bounded linear velocity $v_\text{max}$ and angular velocity $\omega_\text{max}$. At time $t$, the heading deviation required to reach the target at relative position $(x(t), y(t))$ is defined as $\alpha(t)$. The effective reachable radius for this heading is given by


\begin{equation}
r_\text{eff}(t) = v_\text{max} t \cdot \cos\left( \frac{\alpha(t)}{2} \right),
\label{eq:reachable_radius}
\end{equation}

which accounts for the steering needed to reach the target. This reachable area can be visualized as a peanut-shaped cross-section, as seen in Figure \ref{fig:cross-sections}.

The target is considered ``interceptable" whenever the blue curve in Figure \ref{fig:reachability} is within the orange cone. This occurs when the squared Euclidean distance to the predicted position satisfies $\| \mathbf{p}_\text{target}(t) \|^2 \le r_\text{eff}(t)^2$, where $\mathbf{p}_\text{target}(t) = [x(t), y(t)]^\top$ is the target's predicted relative position at time $t$.

\subsubsection{De-coupled Position and Orientation}

When position and orientation are decoupled---e.g. for the spacecraft testbed and the UAV system---we can default to a simpler reachability analysis that uses an effective radius of $r_\text{eff}(t) = v_\text{max} t$, as per Figure \ref{fig:cross-sections}. The procedure is otherwise unaltered.

\subsubsection*{Finding the Interception Point}

Given the observer’s reachable set and the target’s predicted trajectory, we determine the earliest feasible interception point by finding the earliest time at which the target's predicted motion profile pierces the observer’s reachability region (black point in Figure \ref{fig:reachability}). To do this, we solve for the smallest $t^* \in [0, t_\text{max}]$ such that $F(t^*) \triangleq x(t^*)^2 + y(t^*)^2 - r_\text{eff}(t^*)^2  < \delta$, where $\delta$ is a small tolerance. If no solution exists, we set $t = t_\text{max}$ to ensure that interception is still attempted, while providing the maximum allowable time buffer for re-planning.

The final intercept point is $(x^*, y^*) = (x(t^*), y(t^*))$, which is used as the terminal goal for the trajectory planner covered in the next subsection.

\subsection{Observer Trajectory Planning}

This module creates a trajectory of poses from the current observer pose to the predicted interception pose, which is then tracked by the robot's control system. The planning strategy depends on the observer platform, following the same taxonomy as detailed above. For platforms where position and orientation degrees of freedom are coupled, such as the rover, we use a model-based optimization strategy. For platforms where position and orientation are independent, such as the spacecraft testbed and the UAV, we use a model-free optimization strategy.

\subsubsection{Coupled Position and Orientation}
\label{sec:scp}

To generate trajectories when position and orientation are coupled, we solve a receding-horizon, discrete-time sequential convex program (SCP) at each time step.


The decision vector,

\begin{equation}
\mathbf{z} =
\begin{bmatrix}
\mathbf{x}_0^\top & \mathbf{u}_0^\top & \mathbf{x}_1^\top & \mathbf{u}_1^\top & \dots & \mathbf{x}_N^\top
\end{bmatrix}^\top \in \mathbb{R}^{5N+3},
\label{eq:decision_vector}
\end{equation}

\noindent stacks the state and control variables over a horizon of $N$ steps, where $\mathbf{x}_k = [x_k, \ y_k, \ \theta_k]^\top$ and $\mathbf{u}_k = [v_k, \ \omega_k]^\top$ are the states and control inputs, respectively, at time step $k$. Specifically, $(x_k, y_k)$ is the position in the body-frame, $\theta_k$ is the heading, $v_k$ is the forward velocity, and $\omega_k$ is the yaw rate.

The quadratic cost at each SCP iteration consists of stage costs penalizing control magnitudes and deviation from the goal pose, along with terminal costs penalizing final deviation from the goal pose, $\mathbf{x}^\mathrm{goal} = [x_g, \ y_g, \ \theta_g]^\top$. This gives a cumulative cost of


\begin{align}
\!\!\!\!J &= \sum_{k=0}^{N-1} \Bigl[ 
    w_v v_k^2 + w_\omega \omega_k^2 \nonumber \\
&\quad + w_{ip} \left( (x_N - x_g)^2 + (y_N - y_g)^2 \right) + w_{i\theta} (\theta_N - \theta_g)^2 
\Bigr] \nonumber \\
&\quad + w_p \left( (x_N - x_g)^2 + (y_N - y_g)^2 \right)
  + w_\theta (\theta_N - \theta_g)^2,
\label{eq:cost_terms}
\end{align}

\noindent where $w_p$, $w_\theta$, $w_{ip}$, $w_{i\theta}$, $w_v$, and $w_\omega$ are non-negative weights.

In the special case of bidirectional operation, the rover is able to ``back out" of an undesirable state and maneuver into a more favourable state. We call this a \emph{lateral correction}, and frequently encounter such scenarios when the target moves unexpectedly after interception. The optimization problem formulated in this section naturally supports bidirectionality, but an additional bias term can be added to early $v_k$ values to encourage reverse motion first. This is done to break the symmetry of forward-first and reverse-first solutions; it formally prevents us from driving into the target when performing a lateral correction and incentivizes us to back out first.

In each iteration, we impose linearized discrete-time bicycle dynamics by requiring that

\begin{subequations}
    \begin{align}
        \!\!x_{k+1} - x_k &\approx v_k \cos\theta_k^{(i)} \, \Delta t - v_k^{(i)} \sin\theta_k^{(i)} (\theta_k - \theta_k^{(i)}) \Delta t, \\
        \!\!y_{k+1} - y_k &\approx v_k \sin\theta_k^{(i)} \, \Delta t + v_k^{(i)} \cos\theta_k^{(i)} (\theta_k - \theta_k^{(i)}) \Delta t, \\
        \theta_{k+1} &= \theta_k + \omega_k \, \Delta t, \label{eq:dyn_linearized}
    \end{align}
\end{subequations}

where $(\theta_k^{(i)}, v_k^{(i)})$ are values from the current solution about which we're linearizing. Bicycle kinematics are selected because they can generalize to many steering modes, including the one we use, with limited loss in consistency \cite{polack2017}.

We also enforce the initial state condition as $\mathbf{x}_0 = \mathbf{0}$ to encode that we are in the local body-frame and starting from the observer's current pose.

Finally, we limit the vehicle's velocity and turning limits at each iteration using

\begin{align}
-v_\mathrm{max} \le v_k \le v_\mathrm{max}, \quad
-\frac{v_k^{(i)}}{R_\mathrm{min}} \le \omega_k \le \frac{v_k^{(i)}}{R_\mathrm{min}}.
\label{eq:ctrl_limits}
\end{align}

\noindent where $R_\mathrm{min}$ is the minimum turning radius of the observer. This constraint is also linearized about the current solution.

We solve this problem iteratively as a sequential convex program (SCP). At each iteration $i$ of SCP, the quadratic program (QP) is solved to obtain $\mathbf{z}^{(i+1)}$, and the process repeats until $\left\| \mathbf{z}^{(i+1)} - \mathbf{z}^{(i)} \right\| < \varepsilon$ for some tolerance $\varepsilon$ or until a maximum number of SCP iterations is reached. The final $\mathbf{z}^*$ defines the planned trajectory, and is updated online as we make new observations of the target.

\subsubsection{De-coupled Position and Orientation}
\label{sec:min-snap}

We also implement a decoupled polynomial trajectory optimizer for the case where the translational $(x, y, z)$ and yaw ($\psi$) degrees of freedom are independent. The translational motion is generated using the \emph{minimum snap} criterion, while the yaw motion is generated using the \emph{minimum acceleration} criterion, both of which are optimal with respect to control inputs for a UAV \cite{mellinger2011_minsnap}.








This formulation also extends readily to the spacecraft testbed, which has one less degree of freedom but otherwise behaves similarly (since it is holonomic). We simply drop the $z$ component and proceed the same way.

\subsection{Summary}

Algorithm \ref{alg:summary} summarizes the procedural flow of the components described above.

\begin{algorithm}[H]
\caption{Interception Pipeline}
\label{alg:summary}
\begin{algorithmic}[1]
\STATE Initialize EKF state $\mathbf{x}_0$, covariance $\mathbf{P}_0$
\LOOP
    \STATE Acquire AprilTag-based target pose (if available)
    \STATE Estimate target control inputs via EKF state history
    \IF{measurement available}
        \STATE EKF prediction update using observer control inputs and estimated target control inputs
        \STATE EKF measurement update with observed pose
    \ELSE
        \STATE EKF prediction update only
    \ENDIF
    \STATE Predict target trajectory using polynomial regression over last $L$ EKF states (Figure \ref{fig:prediction})
    \STATE Compute interception time $t^*$ and intercept point $(x^*, y^*)$ via reachability analysis (Figure \ref{fig:reachability})
    \IF{observer has coupled position/orientation}
        \STATE Generate observer trajectory by solving an SCP
    \ELSE
        \STATE Generate minimum snap/accel. trajectory
    \ENDIF
    \STATE Apply platform-specific control strategy
\ENDLOOP
\end{algorithmic}
\end{algorithm}

\section{Results and Discussion}
\label{sec:results}

In this section, we present results on all three robotic platforms. We analyze success rates and assess interception errors (both final error and cumulative errors while station-keeping) for a range of target behavior profiles (stochastic and deterministic). We also compare performance between simulation and reality, and conduct physical trials in varying conditions. Finally, we study the effect of our target motion prediction strategy on all platforms when handling sensor dropouts, target occlusions, and partial observability.

\subsection{Results on the Multi-Rover System}

These experiments rely on two rovers. The observer rover is a four-wheel, skid-steered platform measuring 1 m L x 0.79 m W x 0.76 m H (length × width × height) and weighing 120 kg. It features an Nvidia Jetson Orin \cite{jetson_orin} with 12 cores and an Ampere architecture GPU. It's also equipped with a single USB camera that provides a 640 x 480 low-resolution image stream.

The target rover is a dual-body version of the observer, where two four-wheel bodies are connected by an actuator. Each body measures 0.94 m L x 0.66 m W x 0.61 m H and weighs 65 kg. A fiducial is placed on this rover to mark it as the target.

\subsubsection{Overall Interception Accuracy in Simulation and on Real Scenarios}

\begin{figure}[h]
    \centering
    \begin{subfigure}{0.9\linewidth}
        \centering
        \includegraphics[width=\linewidth,trim=0 0 0 45,clip]{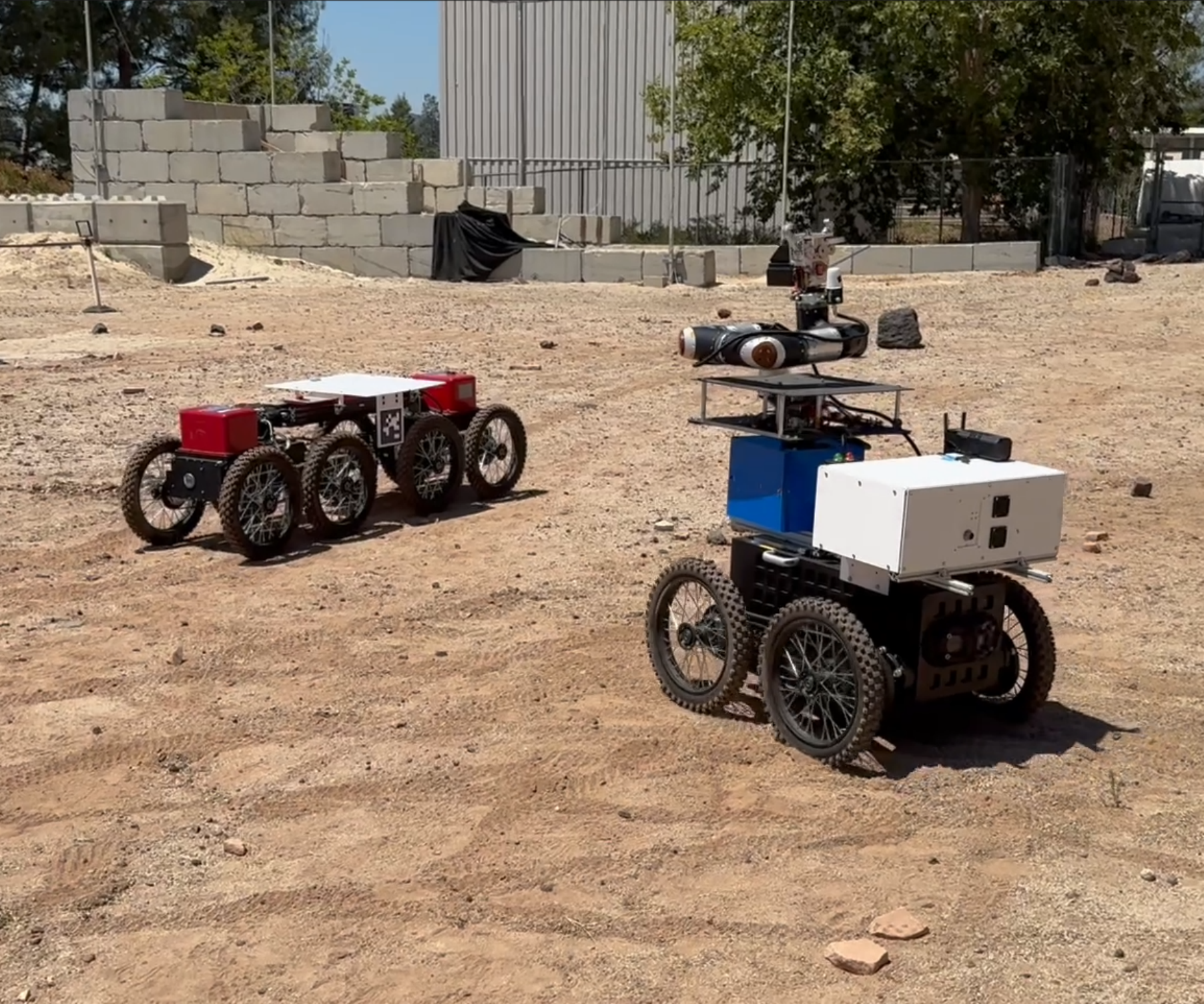}
        \label{fig:rover-examples-a}
    \end{subfigure}
    \caption{An example of the lateral interception case in JPL's Mars Yard. The blue rover is the observer and the red one is the target.}
    \label{fig:rover-examples}
\end{figure}

\begin{figure}[h]
    \centering
    \begin{subfigure}{0.9\linewidth}
        \centering
        \includegraphics[width=\linewidth, trim=40 0 53 0,clip]{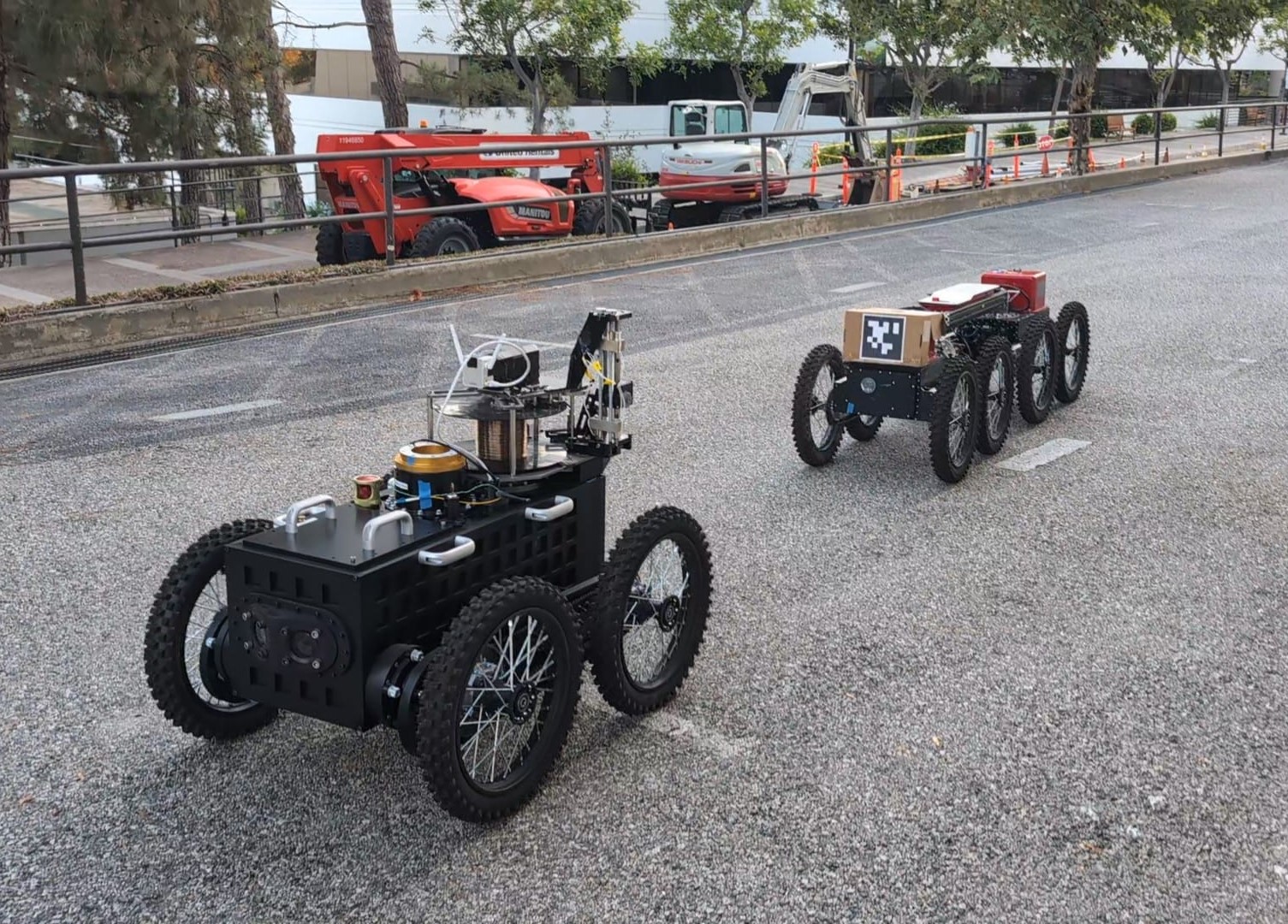}
        \label{fig:rover-examples-b}
    \end{subfigure}
    \hfill
    \begin{subfigure}{0.9\linewidth}
        \centering
        \includegraphics[width=\linewidth, trim=30 30 30 60, clip]{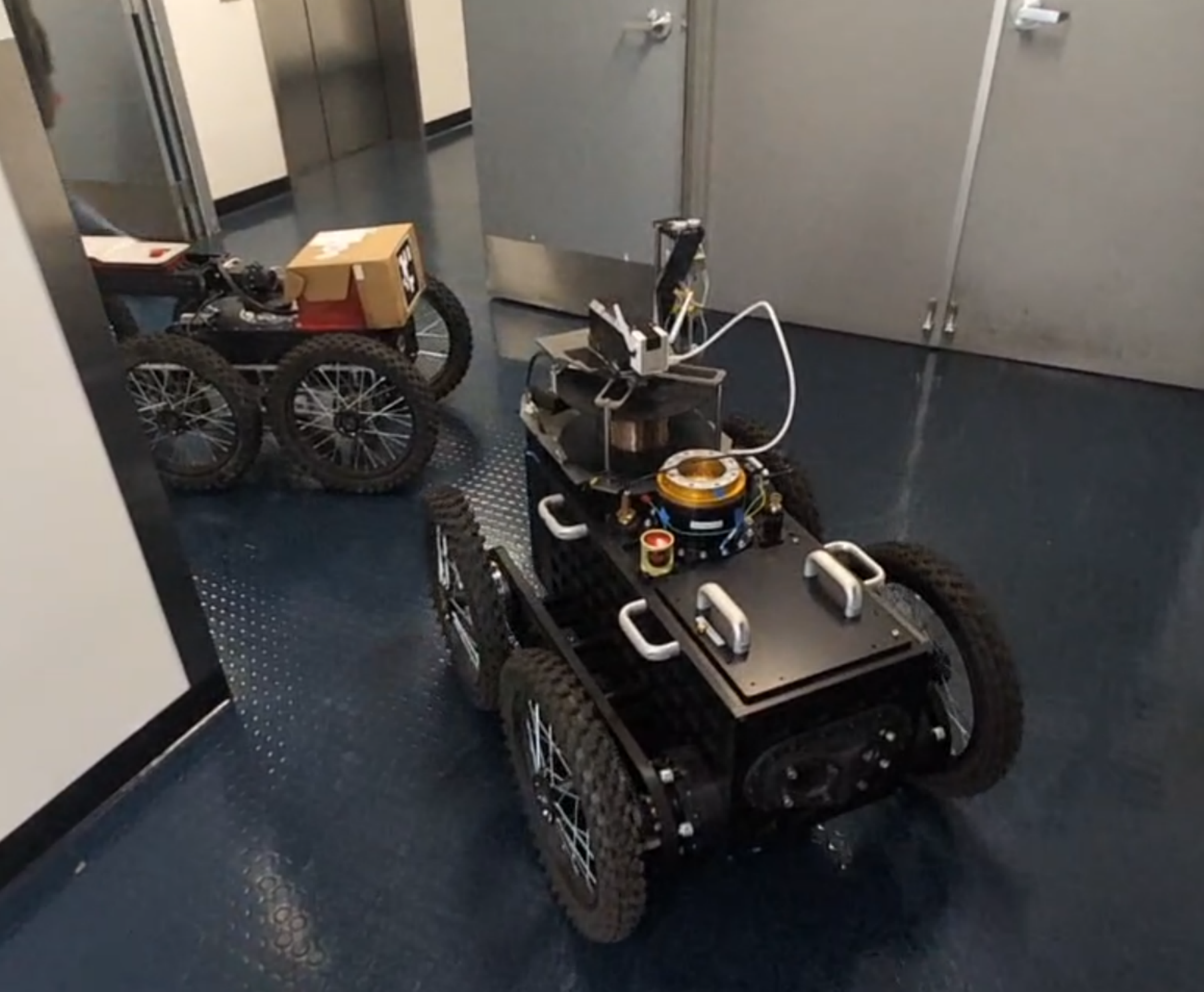}
        \label{fig:rover-examples-b-2}
    \end{subfigure}

    \caption{Two examples of the leader-follower station-keeping case. In both cases, the rover in the foreground is the observer.}
    \label{fig:rover-examples-2}
\end{figure}

To evaluate performance, we conduct systematic experiments in both simulation and on hardware across two types of scenarios: (1) lateral interception, where the target starts several meters away and moves approximately laterally relative to the observer's heading, and (2) leader-follower interception, where the target starts at the same heading as the observer. For case (1), the objective is to achieve a low final interception pose error, while for case (2), the objective is to \emph{maintain} a low pose error \emph{throughout} the scenario, while also ensuring a low final interception pose error. These two cases are illustrated in Figures \ref{fig:rover-examples} and \ref{fig:rover-examples-2} respectively. In all experiments, the observer is commanded to intercept the target at a standoff distance of 1.25 meters, while aligning its orientation as closely as possible with the target.

Hardware experiments are performed on diverse terrain to demonstrate efficacy, including concrete, indoor surfaces, and the natural terrain of JPL’s Mars Yard. Simulation trials are conducted on nominal flat terrain. Results on final pose error for both lateral interception and leader-follower interception are presented in Table \ref{tab:rovers_sim_real}.

\begin{table}[h]
\caption{\textbf{Multi-rover interception performance across different environments and scenarios. *The one failure observed in the Mars Yard was due to a faulty orientation estimate that led to an aggressive maneuver.}}
\centering
\setlength{\tabcolsep}{4pt}
\begin{tabular}{|p{0.16\columnwidth}|p{0.11\columnwidth}|p{0.11\columnwidth}|p{0.11\columnwidth}|p{0.11\columnwidth}|p{0.11\columnwidth}|p{0.11\columnwidth}|}
\hline
\textbf{} & \textbf{Case 1} & \textbf{Case 2} & \textbf{Case 3} & \textbf{Case 4} & \textbf{Case 5} & \textbf{Case 6} \\ \hline
\raggedright\textbf{Type} & Sim. & Sim. & Mars Yard* & Concr-ete & Concr-ete & Indoor \\ \hline
\raggedright\textbf{Scenario} & Lateral & Leader-follower & Lateral & Lateral & Leader-follower & Leader-follower \\ \hline
\hline
\raggedright\textbf{Along-Track\\Error [m]} & 0.016 & 0.143 & 0.044 & 0.075 & 0.106 & 0.048 \\ \hline
\raggedright\textbf{Cross-Track\\Error [m]} & 0.236 & 0.007 & 0.074 & 0.204 & 0.064 & 0.024 \\ \hline
\raggedright\textbf{Angular\\Error [rad]} & 0.041 & 0.077 & 0.158 & 0.115 & 0.135 & 0.190  \\ \hline 
\raggedright\textbf{Successful\\Trials} & 10/10 & 10/10 & 2/3 & 45/45 & 52/52 & 35/35 \\ \hline
\end{tabular}
\label{tab:rovers_sim_real}
\end{table}

We observe consistently strong performance across all scenarios. However, the data reveals a trade-off between final cross-track and angular errors: when one is reduced, the other increases slightly. This reflects the difficulty of our test scenarios, where starting conditions and control limits can prevent simultaneous minimization of both errors.

\subsubsection{Station-Keeping Performance Under Varying Target Motion Profiles}

\begin{figure}[h]
    \centering
    \includegraphics[width=1.0\linewidth]{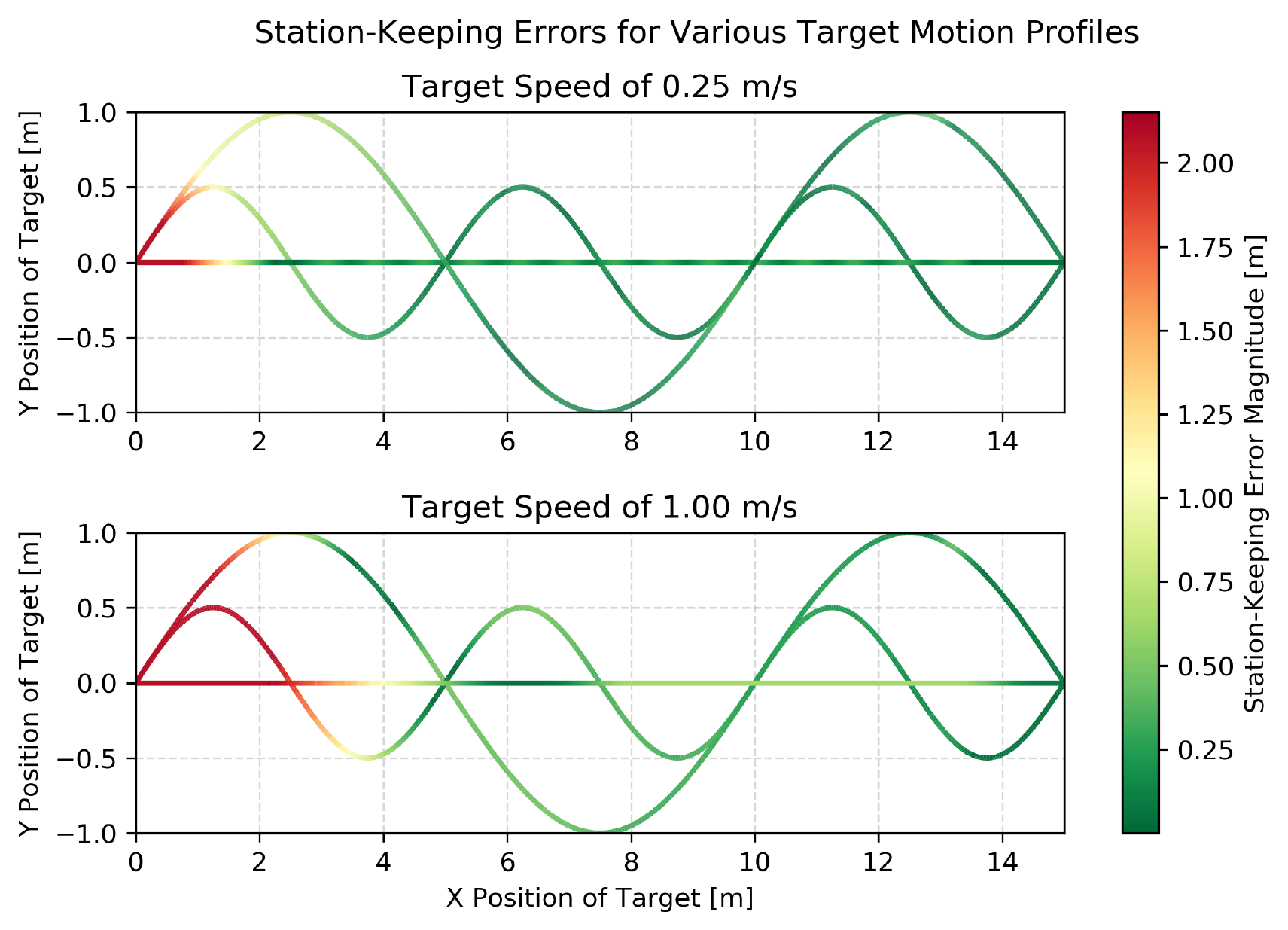}
    \caption{The observer's station-keeping error response to sinusoidal and straight line target motion profiles with varying target speeds.}
    \label{fig:rover_sk}
\end{figure}

Recognizing the small gap between simulation and reality, we leverage simulation for more extensive experimentation on the leader-follower case. We study the performance of our system under various target motion profiles that the observer has no prior knowledge of. We conduct 60 trials, with ten per motion profile. Motion profiles are either sinusoidal (with varying amplitudes, wavelengths, and speeds), or linear (with varying speeds and distance). In these scenarios, the initial pose error is deliberately large, allowing us to assess the system’s ability to correct errors and converge to precise station-keeping, as illustrated in Figure \ref{fig:rover_sk}.

After the rover closes the gap with the non-cooperative leader, we analyze its ability to maintain station. These findings are presented in Table \ref{tab:rovers_sim}. A trial is considered successful if the follower maintains tracking of the target for the full duration and achieves a final interception error, after the target has stopped, below 0.25 meters and 0.20 radians. These thresholds correspond to the maximum empirically determined tolerances for object manipulation (e.g., a physical ‘handshake’) between the two rovers.

\begin{table}[h]
\caption{\textbf{Multi-rover interception and station-keeping performance across varying target speeds, wavelengths, and peak-to-peak motion amplitudes.}}
\centering
\setlength{\tabcolsep}{4pt} 
\begin{tabular}{|p{0.23\columnwidth}|p{0.09\columnwidth}|p{0.09\columnwidth}|p{0.09\columnwidth}|p{0.09\columnwidth}|p{0.09\columnwidth}|p{0.09\columnwidth}|}
\hline
\textbf{} & \textbf{Case 1} & \textbf{Case 2} & \textbf{Case 3} & \textbf{Case 4} & \textbf{Case 5} & \textbf{Case 6} \\ \hline
\raggedright\textbf{Target's Speed [m/s]} & 0.25 & 0.25 & 1.00 & 1.00 & 0.25 & 1.00 \\ \hline
\raggedright\textbf{Sine Wave Wavelength [m]} & 5.0 & 10.0 & 5.0 & 10.0 & N/A & N/A \\ \hline
\raggedright\textbf{Sine Wave\\P2P [m]} & 1.0 & 2.0 & 1.0 & 2.0 & 0.0 & 0.0 \\ \hline
\hline
\raggedright\textbf{Position Error [m]} & 0.071 & 0.080 & 0.043 & 0.095 & 0.080 & 0.056 \\ \hline
\raggedright\textbf{Angular Error [rad]} & 0.089 & 0.093 & 0.130 & 0.104 & 0.004 & 0.012 \\ \hline
\raggedright\textbf{Station-Keeping Pos. Error [m] ± Std.} & 0.342 ± 0.055 & 0.357 ± 0.051 & 0.424 ± 0.188 & 0.436 ± 0.174 & 0.339 ± 0.081 & 0.464 ± 0.231 \\ \hline
\raggedright\textbf{Station-Keeping Ang. Error [rad] ± Std.} & 0.233 ± 0.133 & 0.192 ± 0.111 & 0.207 ± 0.171 & 0.200 ± 0.149 & 0.014 ± 0.020 & 0.030 ± 0.028 \\ \hline
\raggedright\textbf{Successful Trials} & 10/10 & 10/10 & 10/10 & 10/10 & 10/10 & 10/10 \\ \hline
\end{tabular}
\label{tab:rovers_sim}
\end{table}

The results demonstrate a 100\% success rate across all target motion profiles, with consistently low final interception errors. Final position error is largely unaffected by the target’s motion profile which is expected given it depends only weakly on target behavior provided the target is not lost.

Station-keeping errors are notably larger, as expected, but exhibit low variability. It is unreasonable to expect small station-keeping errors in absolute terms, since that would require \emph{a priori} knowledge of the target’s future trajectory. Instead, the low standard deviations highlight the consistency of station-keeping within each scenario, which is our primary metric of interest.

\subsubsection{Analysis of Target Motion Prediction}

In this subsection, we evaluate a scenario where motion prediction is expected to yield a substantial advantage: the target moves with constant velocity but exits the observer’s field of view shortly after the trial begins. In such cases, the system must rely on motion prediction to estimate the target’s location and continue intercepting it. To test this hypothesis, we conduct 15 physical trials with and without motion prediction. Using the same success criteria as in Table \ref{tab:rovers_sim}, the success rate improves dramatically from 6.7\% to 93.3\% when motion prediction is enabled.

\begin{figure}[h]
    \centering
    \includegraphics[width=0.9\linewidth]{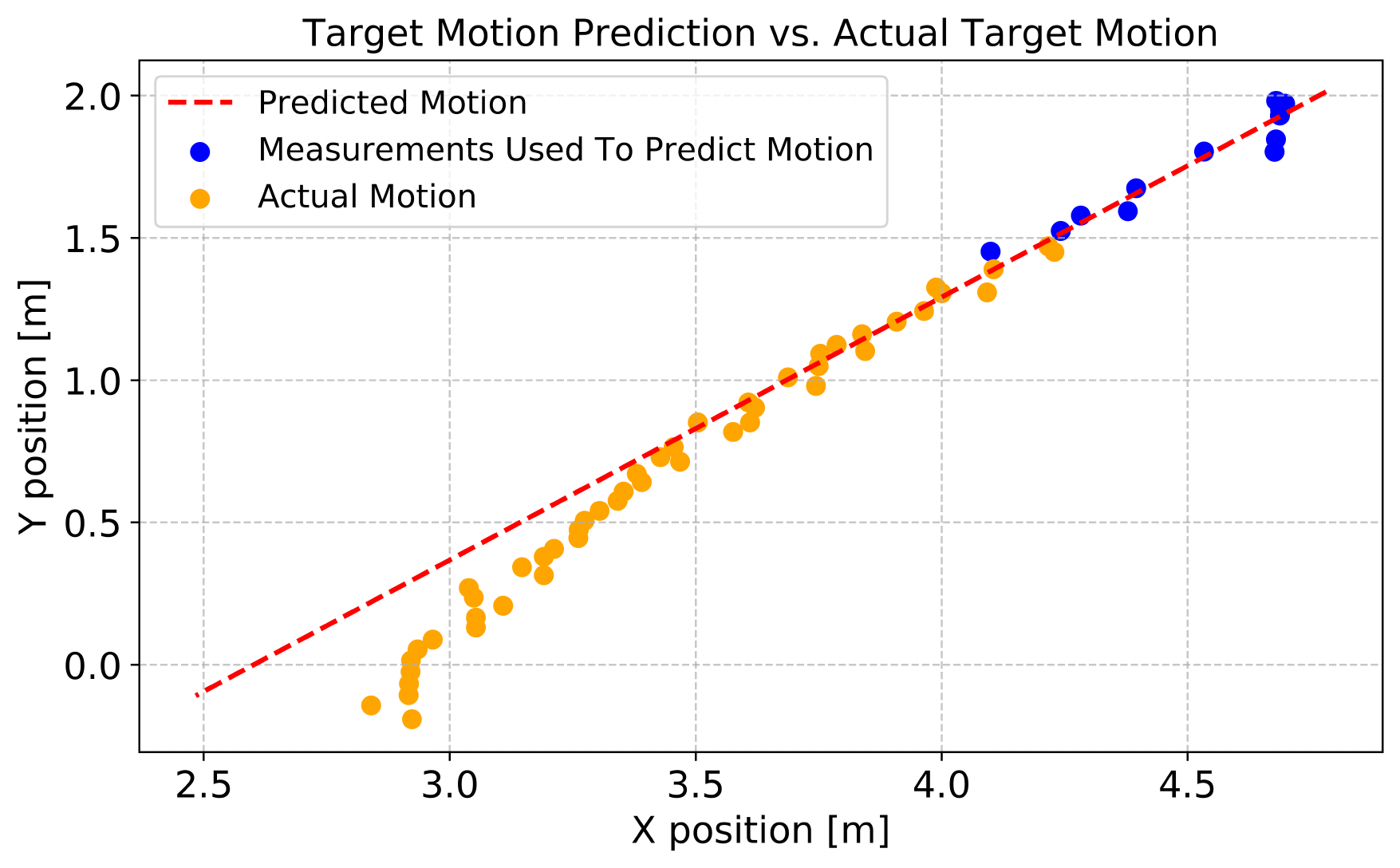}
    \caption{Predicted motion versus actual motion for a target agent. The motion of the target agent starts from the top right corner and ends in the bottom left corner. }
    \label{fig:rover_motion_prediction}
\end{figure}

To understand this improvement, Figure \ref{fig:rover_motion_prediction} compares the actual target trajectory with the predicted trajectory from one of these physical trials. The root mean square error (RMSE) between predicted and actual target position over the predicted horizon is 0.18 meters, demonstrating accurate motion forecasting. This explains the substantial increase in success rate.

We note that for nominal cases with high sensor integrity, motion prediction yields no benefit and can be disabled.

\subsection{Results on the UAV-Target System}

The UAV testbed consists of an X500 quadrotor that employs a PixHawk 6C Mini flight controller and a VOXL2 companion computer. The VOXL2 features 8 cores operating at 3.019 GHz and has 128 GB of flash memory \cite{voxl2}. We use the same low-resolution camera as the rover for perception. To simplify experiments, the target agent is a flat 0.8 m x 0.8 m platform that is pulled by a tether along varying profiles.

\subsubsection{Overall Interception Accuracy in Simulation and in Real Scenarios}

\begin{figure}[h]
    \centering
    \begin{subfigure}{0.49\linewidth}
        \centering
        \includegraphics[width=\linewidth,trim=0 60 0 190,clip]{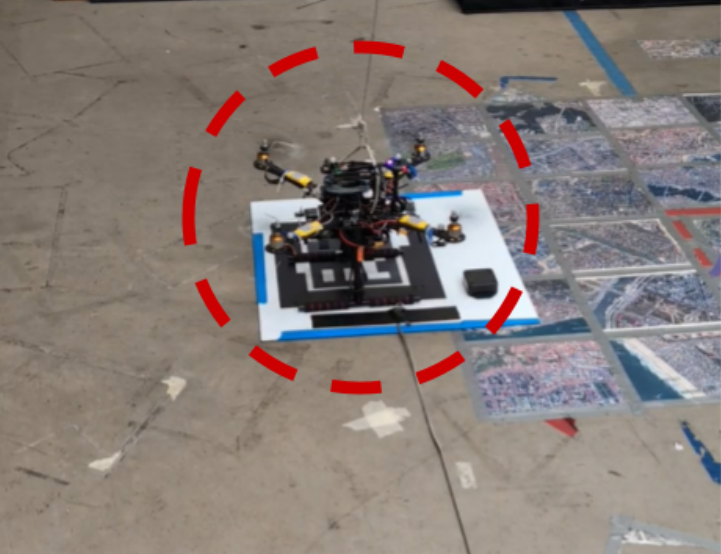}
        \caption{}
        \label{fig:uav-examples-a}
    \end{subfigure}
    \hfill
    \begin{subfigure}{0.49\linewidth}
        \centering
        \includegraphics[width=\linewidth,trim=0 0 0 250,clip]{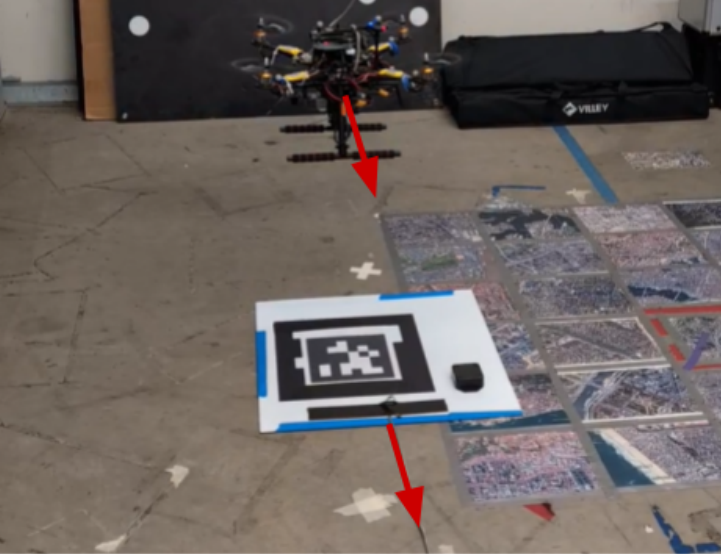}
        \caption{}
        \label{fig:uav-examples-b}
    \end{subfigure}

    \caption{(a) Example of final interception error, measured after landing. (b) Example of station-keeping, where the UAV is instructed to follow the target and align its orientation while maintaining a fixed height above the target.}
    \label{fig:uav-examples}
\end{figure}

We conduct experiments in both simulation and on hardware, considering both static and dynamic targets. For dynamic targets, once the target begins moving, the observer is required to track it from a fixed altitude for a specified duration, and then execute a final landing after a predefined time. Similar to the rovers, this allows us to assess both station-keeping performance and final interception pose error, as illustrated in Figure \ref{fig:uav-examples}. Hardware and simulation results on final pose error are presented in Table \ref{tab:drone_sim_real}.

\begin{table}[h]
\caption{\textbf{UAV interception performance across fixed and dynamic targets in simulation and physical experiments.}}
\centering
\setlength{\tabcolsep}{4pt} 
\begin{tabular}{|p{0.25\columnwidth}|p{0.13\columnwidth}|p{0.13\columnwidth}|p{0.13\columnwidth}|p{0.13\columnwidth}|p{0.13\columnwidth}|}
\hline
\textbf{} & \textbf{Case 1} & \textbf{Case 2} & \textbf{Case 3} & \textbf{Case 4} \\ \hline
\raggedright\textbf{Scenario} & Fixed Target (Sim.) & Dynamic Target (Sim.) & Fixed Target (Real) & Dynamic Target (Real) \\ \hline
\hline
\raggedright\textbf{Position Error [m]} & 0.047 & 0.049 & 0.046 & 0.018 \\ \hline
\raggedright\textbf{Angular Error [rad]} & 0.006 & 0.010 & 0.012 & 0.033 \\ \hline
\raggedright\textbf{Successful Trials} & 10/10 & 10/10 & 3/3 & 3/3 \\ \hline
\raggedright\textbf{Computational Latency [s]} & - & - & 0.053 & 0.051 \\ \hline
\end{tabular}
\label{tab:drone_sim_real}
\end{table}

Similar to the rovers, we observe excellent simulation-to-real transfer. We also see no deterioration in performance when switching to dynamic targets. 


\subsubsection{Station-Keeping Performance Under Varying Target Motion Profiles}

Due to limited access to the testing facility, we perform more extensive studies in a high-fidelity simulator. The simulator uses a differential-drive rover with a 0.8 m × 0.8 m landing platform as the target agent. We evaluate performance under stochastic target motion profiles that the observer has no prior knowledge of, conducting 70 trials—ten per motion profile. Unlike rovers, aerial agents are highly maneuverable and enable experimentation with truly stochastic target motions. At each time step (10 Hz), the target receives random translational and angular velocity inputs within predetermined ranges, ensuring stochastic behavior. For each trial, the target waits for the UAV to take-off, then exhibits stochastic motion for 60 seconds before arriving at a stop. The UAV is required to track the target from a fixed altitude and land on it when it stops. Results are summarized in Table \ref{tab:drone_sim}. A trial is considered successful if the vehicle tracks the target for the full duration and lands in a stable manner (i.e., without falling off after landing).

\begin{table}[h]
\caption{\textbf{UAV interception and station-keeping performance across various translational and angular target velocities.}}
\centering
\setlength{\tabcolsep}{3pt} 
\begin{tabular}{|p{0.24\columnwidth}|p{0.08\columnwidth}|p{0.08\columnwidth}|p{0.08\columnwidth}|p{0.08\columnwidth}|p{0.08\columnwidth}|p{0.08\columnwidth}|p{0.08\columnwidth}|}
\hline
\textbf{} & \textbf{Case 1} & \textbf{Case 2} & \textbf{Case 3} & \textbf{Case 4} & \textbf{Case 5} & \textbf{Case 6} & \textbf{Case 7} \\ \hline
\raggedright\textbf{Target's Transl. Velocities [m/s]} & 0.50 & 1.00 & 0.50 & [0.0, 0.5] & [0.0, 0.5] & [0.0, 1.0] & [0.0, 1.0] \\ \hline
\raggedright\textbf{Target's Angular Velocities [rad/s]} & 0.00 & 0.00 & [-1.0, 1.0] & 0.00 & [-1.0, 1.0] & 0.00 & [-1.0, 1.0] \\ \hline
\hline
\raggedright\textbf{Position Error [m]} & 0.045 & 0.066 & 0.036 & 0.087 & 0.091 & 0.167 & 0.049 \\ \hline
\raggedright\textbf{Angular Error [rad]} & 0.007 & 0.008 & 0.030 & 0.244 & 0.151 & 0.051 & 0.042 \\ \hline
\raggedright\textbf{Station-Keeping Pos. Error [m] ± Std.} & 0.510 ± 0.066 & 0.971 ± 0.177 & 0.679 ± 0.206 & 0.392 ± 0.051 & 0.493 ± 0.181 & 0.654 ± 0.187 & 0.496 ± 0.134 \\ \hline
\raggedright\textbf{Station-Keeping Ang. Error [rad] ± Std.} & 0.006 ± 0.007 & 0.008 ± 0.008 & 0.159 ± 0.163 & 0.005 ± 0.006 & 0.096 ± 0.067 & 0.006 ± 0.005 & 0.084 ± 0.058 \\ \hline
\raggedright\textbf{Successful Trials} & 10/10 & 7/10 & 9/10 & 10/10 & 9/10 & 9/10 & 9/10 \\ \hline
\end{tabular}
\label{tab:drone_sim}
\end{table}

Results are strong, though some notable failures occur. The lowest success rate (70\%) and largest station-keeping position errors arise in case 2, where the target moves at a constant speed in a straight line. This counterintuitive outcome stems from the fact that the observer’s maximum speed is constrained to match that of the target, making it difficult to maintain station when the target moves steadily at that limit. In contrast, when the target’s velocity is stochastic, its average speed is lower, allowing the observer to perform rate-matching and keep station more effectively.

Also, as expected, the highest mean and variance in station-keeping orientation error occur when stochastic angular velocities are introduced, though these errors remain bounded below 0.16 radians.

\subsubsection{Utility of Target Motion Prediction Under Sensor Dropout}

\begin{figure}[h]
    \centering
    \includegraphics[width=1.0\linewidth]{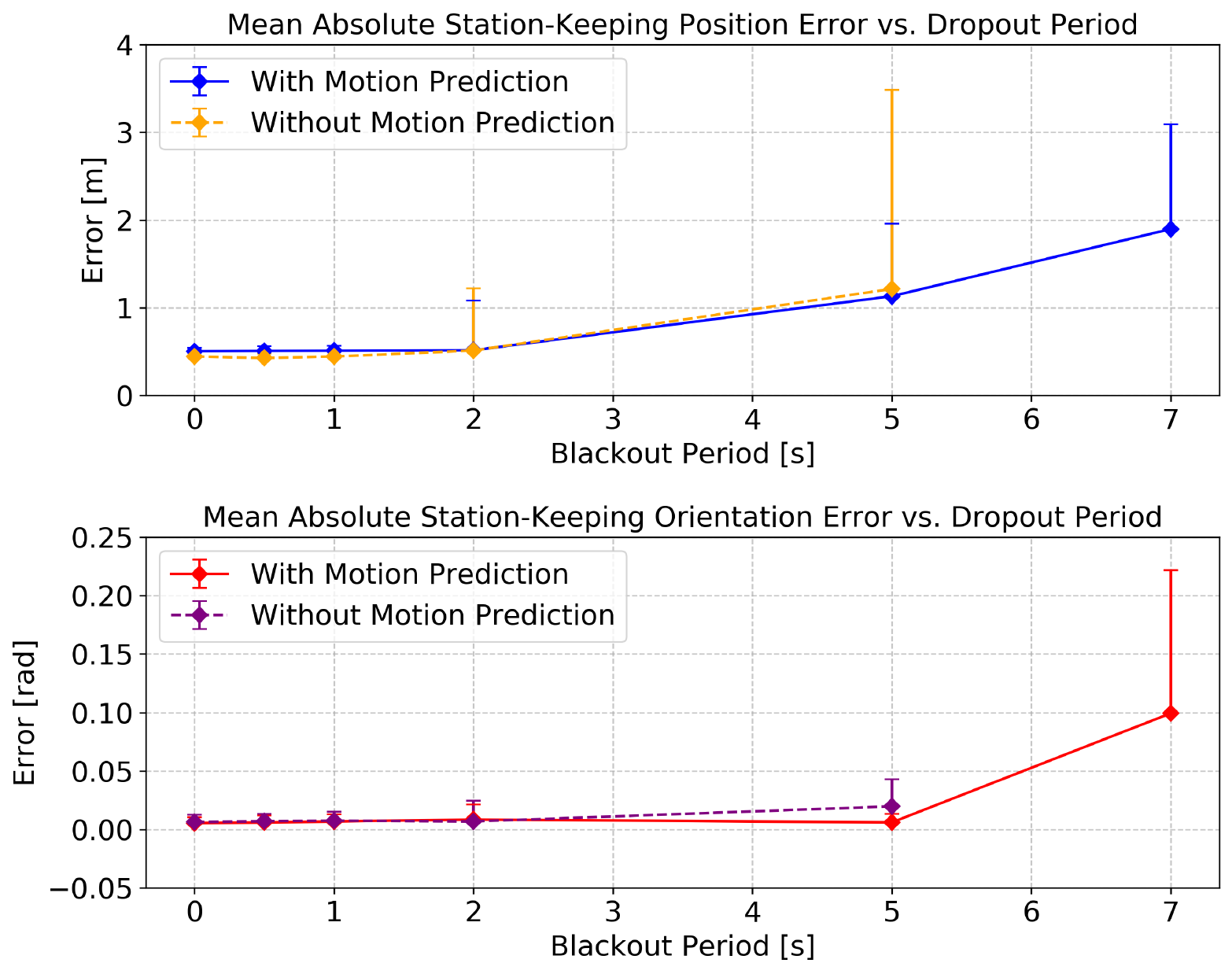}
    \caption{Mean position and orientation station-keeping errors, along with their standard deviations for various sensor dropout periods with and without motion prediction. Only positive error bars are shown since the plot presents mean absolute errors.}
    \label{fig:drone_dropout_errors}
\end{figure}

Our system leverages target motion prediction to handle extended sensor dropouts. We evaluate this capability by simulating dropouts of varying durations, triggered repeatedly with 5-second intervals throughout each 60-second trial. To ensure reproducibility, the target follows a deterministic linear trajectory at 0.5 m/s. We compare success rates and station-keeping performance with and without motion prediction. Results are summarized in Figure \ref{fig:drone_dropout_errors}.


Figure \ref{fig:drone_dropout_errors} shows a modest but critical improvement in success rates with motion prediction, as the system fails completely without it for dropouts longer than 7 seconds. The figure also reveals that motion prediction actually does \emph{not} reduce the mean station-keeping error—--and can even slightly increase it---but it substantially lowers the standard deviation of the error during long sensor dropouts, indicating more \emph{consistent} performance. In other words, while motion prediction may slightly degrade nominal performance, it becomes a key advantage under severe sensor dropout conditions.

\subsection{Results on the Air-Thruster Spacecraft Testbed}

Due to current hardware limitations, only simulation results are presented for the spacecraft testbed at this time. While the software has been successfully deployed on the testbed and has consistently demonstrated reliable dynamic station-keeping, the system is not yet prepared for comprehensive experimental validation. Results from physical testing will be included in future work or in the final version of this paper.

\subsubsection{Station-Keeping Performance Under Varying Target Motion Profiles}

Table \ref{tab:spacecraft_stationkeeping} summarizes station-keeping performance under various stochastic target motion profiles. Both the observer and target are highly maneuverable. To emulate realistic stochastic behavior, the target is driven by randomly sampled control inputs (translational and angular velocities), similar to the UAV platform.

\begin{table}[h]
\caption{\textbf{Air-thruster spacecraft testbed station-keeping performance across various translational and angular target velocities.}}
\centering
\setlength{\tabcolsep}{3pt} 
\begin{tabular}{|p{0.32\columnwidth}|p{0.14\columnwidth}|p{0.14\columnwidth}|p{0.14\columnwidth}|p{0.14\columnwidth}|}
\hline
\textbf{} & \textbf{Case 1} & \textbf{Case 2} & \textbf{Case 3} & \textbf{Case 4} \\ \hline
\raggedright\textbf{Target's Transl. Velocities [m/s]} & 0.1 & 0.1 & [0.0, 0.1] & [0.0, 0.1] \\ \hline
\raggedright\textbf{Target's Angular Velocities [rad/s]} & 0.0 & [-1.0, 1.0] & 0.0 & [-1.0, 1.0] \\ \hline
 \hline
\raggedright\textbf{Position Error [m]} & 0.023 & 0.028 & 0.021 & 0.024 \\ \hline
\raggedright\textbf{Angular Error [rad]} & 0.019 & 0.022 & 0.010 & 0.030 \\ \hline
\raggedright\textbf{Station-Keeping Pos. Error [m] ± Std.} & 0.034 ±
0.020 & 0.042 ± 0.022 & 0.025 0.017 & 0.024 0.018 \\ \hline
\raggedright\textbf{Station-Keeping Ang. Error [rad] ± Std.} & 0.236 ± 0.076 & 0.274 ± 0.096 & 0.088 ± 0.044 & 0.163 ± 0.133\\ \hline
\raggedright\textbf{Successful Trials} & 10/10 & 10/10 & 10/10 & 10/10 \\ \hline
\end{tabular}
\label{tab:spacecraft_stationkeeping}
\end{table}

Station-keeping performance for both position and orientation improves when the target’s speed is stochastic rather than constant. This reflects the same underlying phenomenon observed for the UAV: when the target moves at a constant speed equal to the observer’s maximum, station-keeping becomes challenging. In contrast, stochastic velocities yield a lower effective average speed, which enables more reliable station-keeping.

\subsubsection{Utility of Target Motion Prediction Under Sensor Dropout}

\begin{figure}[h]
    \centering
    \includegraphics[width=\linewidth]{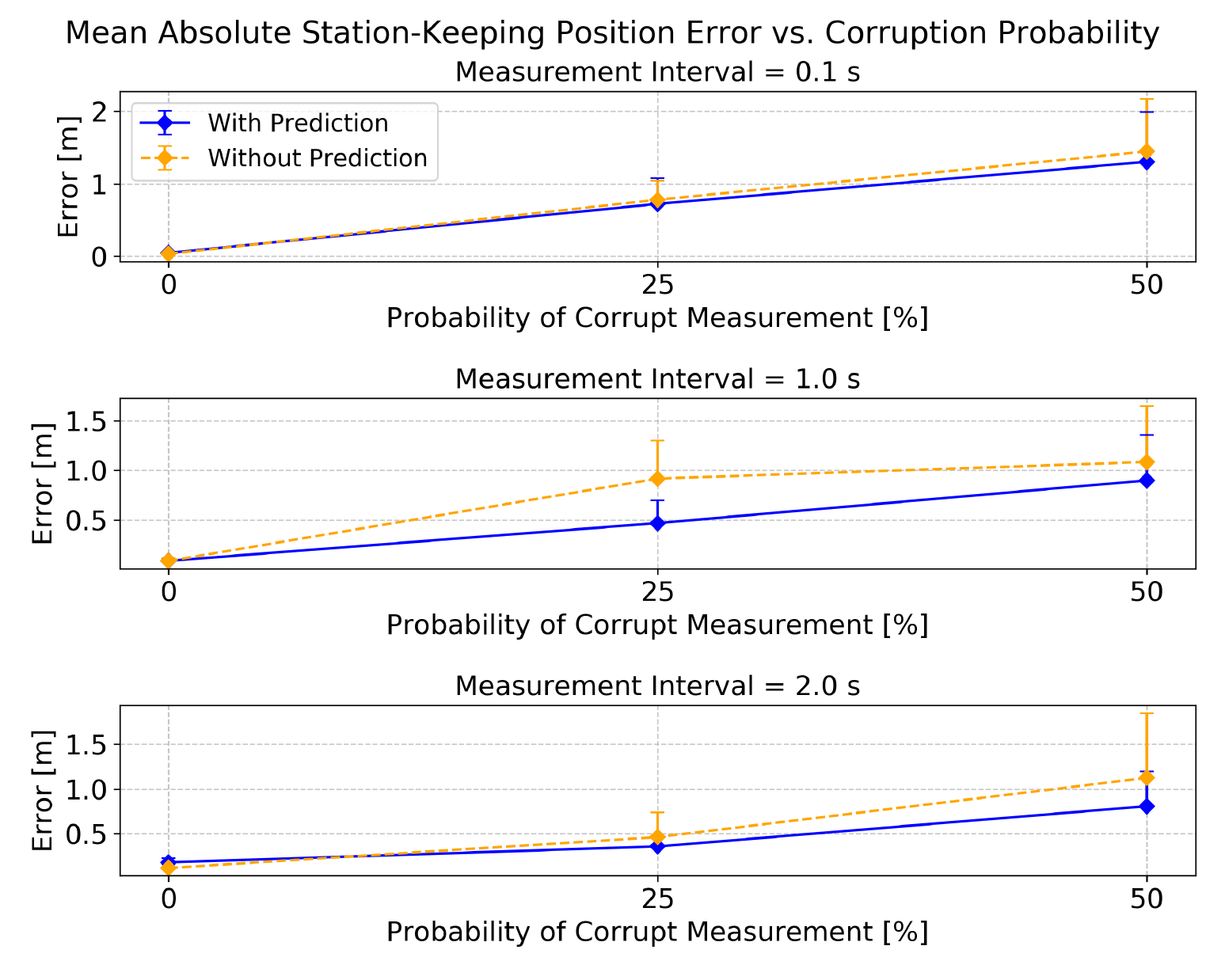}
    \caption{Mean absolute station-keeping position error and standard deviation of the air-thruster spacecraft testbed under varying sensor conditions (measurement dropouts and corrupted observations). Only positive error bars are shown since the plot presents mean absolute errors.}
    \label{fig:spacecraft_sensor_dropout}
\end{figure}

This platform also enables evaluation under varying sensor conditions. Figure \ref{fig:spacecraft_sensor_dropout} shows mean station-keeping position error with different time delays between measurements and varying probabilities of receiving corrupted observations. Corrupt observations are simulated by injecting fully stochastic range and bearing readings with no correlation to the actual range and bearing.

Figure \ref{fig:spacecraft_sensor_dropout} agrees with the other two robotic platforms in that motion prediction yields a significant advantage for off-nominal cases (in this case, when the probability of receiving corrupt measurements is high), but offers little improvement under nominal operating conditions. Interestingly, the results indicate that high measurement frequencies can actually degrade performance when measurements are highly corrupted. This is because frequent low-quality measurements are more detrimental than infrequent low-quality measurements.

\section{Conclusions and Future Work}
\label{sec:conc}

This work presents a simple yet effective general-purpose framework for robotic interception of dynamic, non-cooperative targets using only monocular vision in the observer body-frame. By integrating an EKF-based relative pose estimator, a history-conditioned target motion predictor, and a real-time trajectory planner, the proposed method achieves reliable interception across heterogeneous platforms, including UAVs, ground rovers, and air-thruster spacecraft simulators.

Extensive experiments demonstrate centimeter-level interception accuracy, high success rates under both deterministic and stochastic target motion, and real-time performance in both simulation and on embedded hardware. Results also study the effect of diverse partial observability conditions and assess our proposed mitigation strategies. Overall, results highlight the robustness, versatility, and computational efficiency of our approach, even under limited sensing and without any global localization.

Future work will extend the framework to leverage diffusion-based intent prediction for highly erratic targets and will conduct physical trials on the spacecraft testbed. Learning-based target detection strategies developed at JPL will be incorporated to address off-nominal image quality, including a formal comparison against classical methods. We also intend to formally handle both static and dynamic obstacles in our planning module to further enhance robustness.

\section*{Acknowledgements}

We thank Roland Brockers for his guidance during UAV development and testing; Gerik Kubiak for his support in UAV integration and testing; Katherine Tighe for assistance with 3D printing; Ben Cheng for supervising rover data collection; Abhay Negi for helping with data collection and UAV mechanical design; Timothy Lui for integrating the project onto the spacecraft testbed; Shelby Hackett for support in using the testbed; and Kevin Lo for supervision of the testbed. The research was carried out at the Jet Propulsion Laboratory, California Institute of Technology, under a contract with the National Aeronautics and Space Administration (80NM0018D0004).

\bibliographystyle{IEEEtran}
\bibliography{references}

\thebiography
\begin{biographywithpic}
{Tanmay Patel}{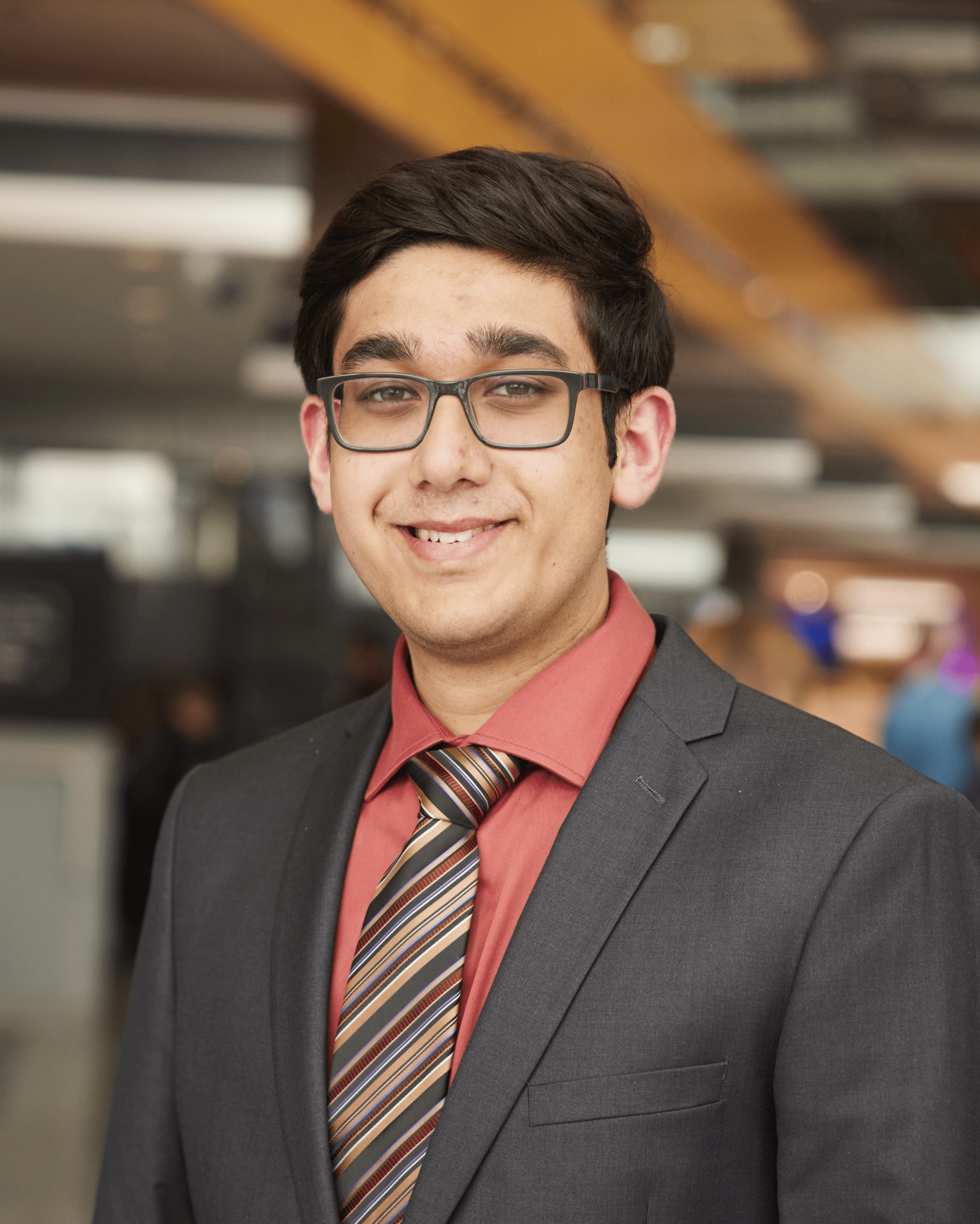}
is an undergraduate student at the University of Toronto, pursuing a BASc in Engineering Science with a major in Robotics and a minor in AI. He has conducted research in healthcare informatics at the Centre for Addiction and Mental Health (Toronto), in multi-agent robotics at the Technical University of Munich, and in autonomous urban driving at the University of Toronto Institute for Aerospace Studies. Most recently, he was a research intern at the NASA Jet Propulsion Laboratory where he completed this work.
\end{biographywithpic} 

\begin{biographywithpic}
{Erica Tevere}{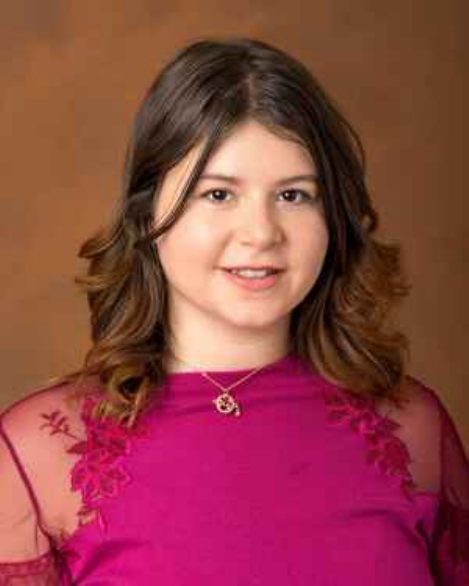}
is a Robotics Technologist in the robotics mobility group at NASA Jet Propulsion Laboratory. She received her Master's in Robotics from Johns Hopkins University and her Bachelor's of Mechanical Engineering from the University of Michigan. She has supported projects for NASA, DARPA, and other DOD organizations on developing and deploying autonomous solutions to fielded systems. Her research interests focus primarily on autonomy for extreme environment robots with specific interests in visually degraded and resource constrained environments.
\end{biographywithpic}

\begin{biographywithpic}
{Erik Kramer}{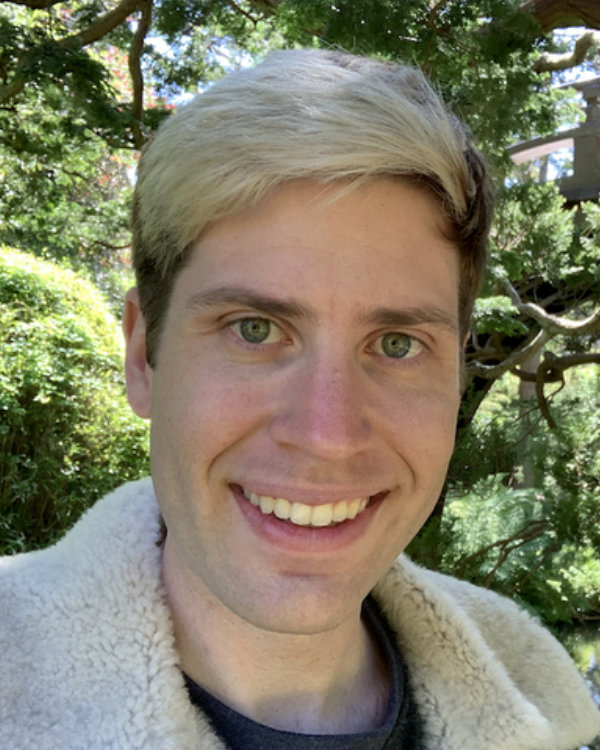}
is a Postdoctoral Fellow at NASA Jet Propulsion Laboratory in the Maritime And Multi-Agent Autonomy group. His doctoral work focused on hardware and software development of collaborative multi-arm robotic exoskeletons for virtual reality powered stroke rehabilitation. At JPL he has developed algorithms for an Ocean World lander robotic arm testbed as well as served as the responsible engineer for in-space servicing, assembly, and manufacturing (ISAM) testbed technology development and operation. He received a BA in Physics and BS in Mechanical Engineering from the University of California, Berkeley and a MS and PhD in Mechanical Engineering from the University of California, Los Angeles (UCLA).
\end{biographywithpic}

\begin{biographywithpic}
{Rudranarayan Mukherjee}{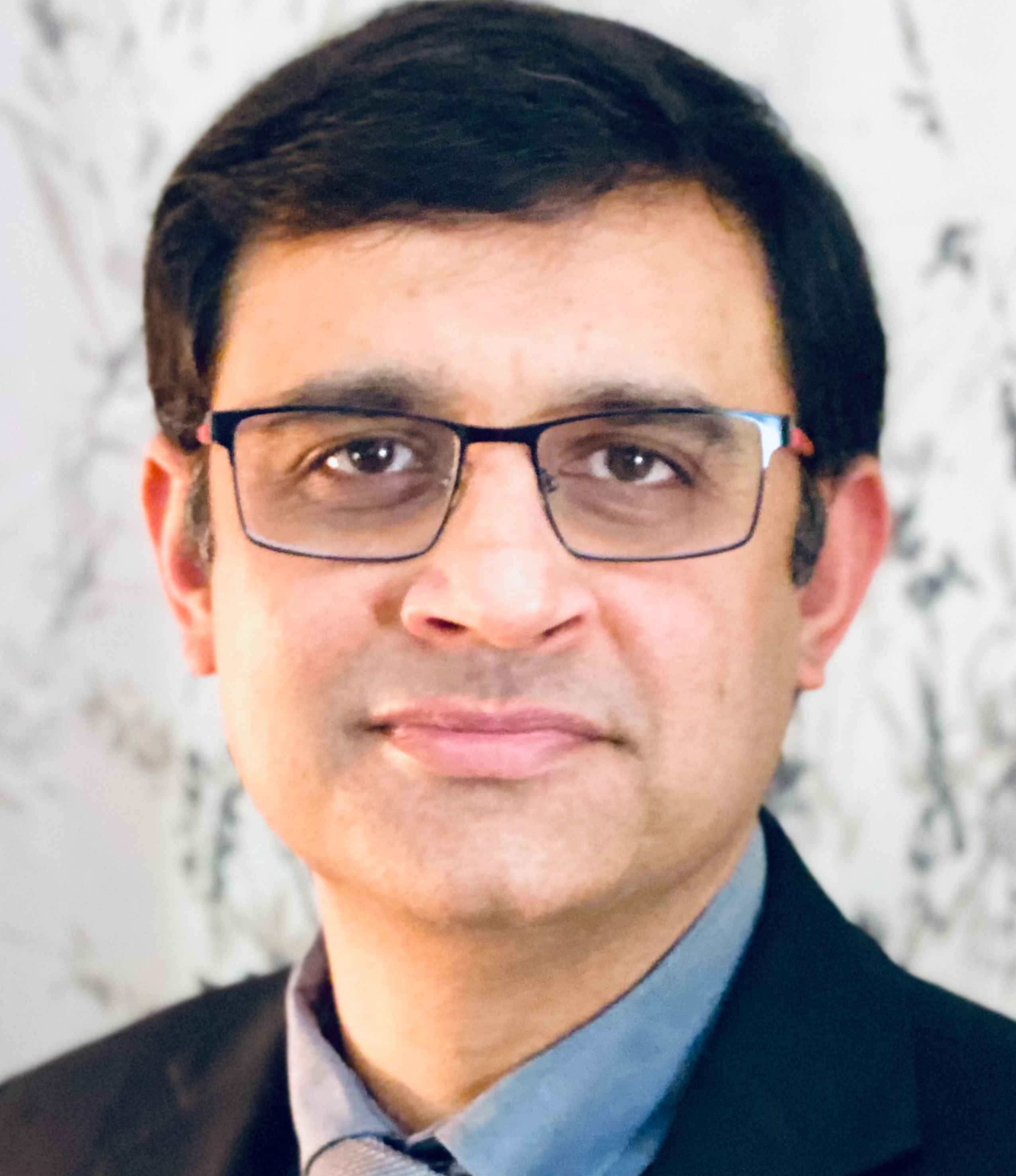}
is the JPL institutional Lead for In-space Servicing, Assembly, and Manufacturing (ISAM) and principal investigator for various robotics efforts. He holds a PhD in Mechanical Engineering. He develops technologies for space and terrestrial applications to demonstrate the art of possible and seed new missions or application areas.
\end{biographywithpic}

\end{document}